%% file: main.tex
\documentclass{article}

\PassOptionsToPackage{numbers, compress}{natbib}

    \usepackage[preprint]{neurips_2025}

\usepackage[utf8]{inputenc} 
\usepackage[T1]{fontenc}    
\usepackage{hyperref}       
\usepackage{url}            
\usepackage{graphicx}        
\usepackage{booktabs}        
\usepackage[table,xcdraw]{xcolor}  
\usepackage{makecell}        
\usepackage{amssymb}         
\usepackage{textcomp}        
\usepackage{multirow}        
\usepackage{pifont}

\input{math_commands.tex}

\usepackage{mathtools}
\usepackage{graphicx}
\usepackage{float}
\usepackage{color}
\usepackage{booktabs}    
\usepackage{colortbl}    
\usepackage[table,xcdraw]{xcolor}  
\usepackage{amssymb}     
\usepackage{graphicx}    

\usepackage{bm}
\usepackage{wasysym}
\usepackage{amsmath}
\usepackage{amssymb}
\usepackage{amsfonts}
\usepackage{multirow}
\usepackage{adjustbox}
\usepackage{booktabs}
\usepackage{xcolor,colortbl}
\usepackage{caption}
\usepackage{subcaption}
\usepackage{wrapfig}
\usepackage{nicefrac}

\usepackage{booktabs}
\usepackage{algorithm}
\usepackage{algpseudocode}
\usepackage{listings}
\usepackage{tikz}

\usepackage{amssymb,amsmath,amsthm}
\newcommand{\cmark}{\ding{51}}  
\newcommand{\xmark}{\ding{55}}  

\newcommand*{\belowrulesepcolor}[1]{%
  \noalign{%
    \kern-\belowrulesep 
    \begingroup 
      \color{#1}%
      \hrule height\belowrulesep 
    \endgroup 
    \vspace{-0.03mm}
  }%
} 
\newcommand*{\aboverulesepcolor}[1]{%
  \noalign{%
  \vspace{-0.03mm}
    \begingroup 
      \color{#1}%
    \endgroup 
    \kern-\aboverulesep 
  }%
}
\title{T2UE: Generating Unlearnable Examples from Text Descriptions}

\author{%
  Xingjun Ma \\
  Fudan University\\
  Shanghai, China\\
  \texttt{xingjunma@fudan.edu.cn} \\
  \And
  Hanxun Huang \\
  The University of Melbourne \\
  Melbourne, Australia \\
  \texttt{hanxun@unimelb.edu.au} \\
  \AND
  Tianwei Song \\
  Fudan University \\
  Shanghai, China \\
  \texttt{22210240266@m.fudan.edu.cn} \\
  \And
  Ye Sun \\
  Fudan University \\
  Shanghai, China \\
  \texttt{yesun23@m.fudan.edu.cn} \\
  \And
  Yifeng Gao \\
  Fudan University \\
  Shanghai, China \\
  \texttt{yifenggao23@m.fudan.edu.cn} \\
  \And
  Yu-Gang Jiang\thanks{Corresponding author} \\
  Fudan University \\
  Shanghai, China \\
  \texttt{ygj@fudan.edu.cn} \\
}

\begin{document}

\maketitle

\begin{abstract}
Large-scale pre-training frameworks like CLIP have revolutionized multimodal learning, but their reliance on web-scraped datasets, frequently containing private user data—raises serious concerns about misuse. Unlearnable Examples (UEs) have emerged as a promising countermeasure against unauthorized model training, employing carefully crafted unlearnable noise to disrupt the learning of meaningful representations from protected data.
Current approaches typically generate UEs by jointly optimizing unlearnable noise for both images and their associated text descriptions (or labels). However, this optimization process is often computationally prohibitive for on-device execution, forcing reliance on external third-party services. This creates a fundamental privacy paradox: users must initially expose their data to these very services to achieve protection, thereby compromising privacy in the process. Such a contradiction has severely hindered the development of practical, scalable data protection solutions.
To resolve this paradox, we introduce \textbf{Text-to-Unlearnable Example (T2UE)}, a novel framework that enables users to generate UEs using only text descriptions. T2UE circumvents the need for original image data by employing a text-to-image (T2I) model to map text descriptions into the image (noise) space, combined with an error-minimization framework to produce effective unlearnable noise. Extensive experiments show that T2UE-protected data substantially degrades performance in downstream tasks (e.g., cross-modal retrieval) for state-of-the-art models. Notably, the protective effect generalizes across diverse architectures and even to supervised learning settings. Our work demonstrates the feasibility of ``zero-contact data protection", where personal data can be safeguarded based solely on their textual descriptions, eliminating the need for direct data exposure.
\end{abstract}

\section{Introduction}
The advent of large-scale multi-modal models, such as CLIP \cite{radford2021learning} and ALIGN \cite{jia2021scaling}, has significantly advanced multi-modal pre-training by demonstrating powerful capabilities in bridging vision and language.
These models derive their strength from pre-training on massive, non-curated web-scraped datasets, often comprising billions of image-text pairs \cite{schuhmann2021laion, schuhmann2022laion, gadre2023datacomp, xu2024demystifying}.
A growing concern is that several large-scale datasets have been collected without user consent and subsequently used to train commercial models \cite{birhane2021large, hill2022secretive, longpre2023data}.
The potential for unauthorized exploitation of such data for model training underscores an urgent need for effective data protection methods.

\begin{figure*}[t]  
    \centering  
    \includegraphics[width=0.9\textwidth]{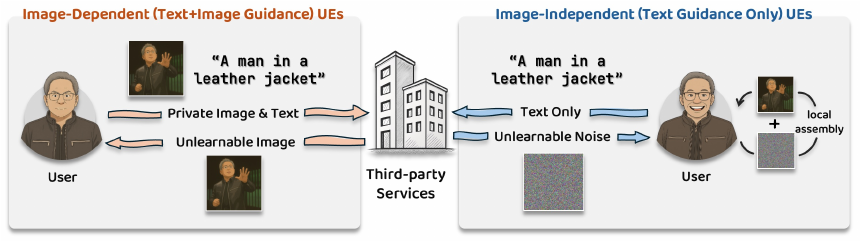}
    \caption{Illustration of our T2UE data protection approach. Existing image-dependent methods require access to both sensitive images and their corresponding text for third-party generation of UE, introducing a risk of data leakage. In contrast, image-independent T2UE shares only textual descriptions with the third party, ensuring \textit{zero contact} with the sensitive image.} 
    \vspace{-0.11in}
    \label{fig:intro_t2ue}
\end{figure*}

Unlearnable Examples (UEs) have emerged as a promising proactive strategy for data misuse preservation \cite{huang2021unlearnable,fu2022robust,ren2022transferable,zhang2023unlearnable,sadasivan2023cuda,wang2024unlearnable,liu2024stable,wang2024provably,ye2025how,gong2025armor,ma2025safety}. By introducing meticulously crafted, often imperceptible perturbations into data, UEs aim to disrupt the training process, rendering the data unusable for effective model learning. The underlying mechanism often involves creating misleading features that the model easily learns instead of the true data features \cite{sandoval2023can}. While initially developed for uni-modal tasks like image classification \cite{huang2021unlearnable, fowl2021adversarial} and subsequently extended to areas like segmentation \cite{sun2024unseg}, 3D point clouds \cite{wang2024unlearnable}, time series \cite{jiang2024unlearnable}, and neural code learning \cite{ji2022unlearnable}.
Adapting UEs to the multi-modal domain and protecting against large-scale pre-training presents unique challenges, as it requires disrupting the learned alignment between continuous image features and diverse textual semantics, rather than merely associating perturbations with discrete class labels \cite{huang2021unlearnable}.

The existing approach for multi-modal unlearnable examples (MUEs) exemplified by multi-step error minimization (MEM) \cite{liu2024multimodal}, proposes jointly optimizing image perturbations alongside specific textual "triggers" appended to the original captions. However, this strategy exhibits significant limitations hindering practical adoption. Firstly, its reliance on discrete textual triggers makes the unlearnable effect fragile and can be easily nullified if the triggers are altered, paraphrased, or removed by common text processing, undermining robustness. Optimizing these triggers is also computationally inefficient. 
Second, and more critically, MEM as well as most existing UE methods, requires access to the original user image to compute perturbations. This process is often computationally intensive and may depend on hardware or resources not readily available to end users, necessitating the use of third-party services. As illustrated in Figure \ref{fig:intro_t2ue}, this creates a fundamental paradox:~\textit{To safeguard data, users must first expose their potentially sensitive data (e.g. by uploading them to an external service), creating inherent risks of stealing, intercepting, misuse, or leakage during the protection process itself. }

In this work, we introduce \textbf{Text-to-Unlearnable Example (T2UE)}, a novel paradigm that fully decouples the generation of unlearnable examples (UEs) from source image data. As illustrated in Figure \ref{fig:intro_t2ue}, our key innovation lies in learning a mapping from the textual semantic space to the unlearnable perturbation space, enabling the creation of effective protective perturbations guided solely by textual descriptions. This approach eliminates the need for original image access during the protection phase, reducing the risk of data exposure. The perturbations produced by T2UE are semantically tied to the text and are specifically designed to disrupt the contrastive alignment objective central to multi-modal pre-training, regardless of image content. Furthermore, T2UE improves task transferability, a known limitation of prior UE methods. By leveraging a pre-trained CLIP model as a surrogate to simulate the outer optimization in the error-minimization framework \cite{huang2021unlearnable}, we show that T2UE-generated UE generalize beyond contrastive learning to impact supervised tasks. Collectively, these contributions mark a significant step toward practical data misuse protection in large-scale multi-modal pre-training.

In summary, our main contributions are:
\begin{itemize}
    \item We propose T2UE, a novel UE generation framework that operates solely on textual input, overcoming the critical \textit{image-dependency} limitation of existing approaches.

    \item We introduce a simple yet powerful approach to achieve T2UE. We use a text-guided unlearnable generation model to learn a mapping from the text semantic space to the unlearnable perturbation space, by leveraging CLIP encoders as a surrogate model. 

    \item We conduct comprehensive evaluations demonstrating that T2UE effectively prevents data misuse in contrastive pre-training, exhibits strong protection transferability across diverse multi-modal architectures and tasks, as well robust to different scenarios. 

\end{itemize}

\section{Related Work}

\noindent\textbf{Unlearnable Examples.} Unlearnable Examples (UEs) represent a proactive privacy-preserving technique that prevents effective learning by machine learning models through the injection of carefully crafted, imperceptible noise into training data~\cite{huang2021unlearnable}. Subsequent research elucidated their mechanism as the creation of easily learnable input-output "shortcuts" that mislead model training~\cite{sandoval2023can, sandoval2022autoregressive}. Enhancing practicality has driven efforts to improve UE robustness against defenses like adversarial training~\cite{yang2024robust, liu2024stable} and boost transferability across learning paradigms and tasks~\cite{huang2021unlearnable}. Notably, UE applications have expanded beyond classification to finer-grained tasks like image segmentation (e.g., UnSeg~\cite{sun2024unseg}) and are emerging for other modalities like 3D point clouds~\cite{wang2024unlearnable}. 
However, extending UEs to multimodal models like CLIP is challenging; existing methods often jointly optimize image and text perturbations~\cite{liu2024multimodal}, facing issues with text optimization difficulty and inherent privacy risks from requiring original image access. Indeed, as summarized in Table~\ref{tab:method_overview}, this critical requirement for image access is common among most prior UE techniques and the existing MUE method MEM, starkly contrasting with the image-independent approach of our proposed T2UE. Distinct from prior work, our method generates protective noise solely using textual information. This text-only approach circumvents the limitations of previous MUE strategies, offering a more secure, robust, and practical pathway for multimodal data privacy.

\begin{table*}[t]
\centering 
  \caption{A summary of existing UE generation methods. Properties fully satisfied (\cmark) and not satisfied (\xmark) are indicated for each method.}
  \vspace{-0.4cm}
\label{tab:method_overview}
\begin{adjustbox}{width=0.9\linewidth}
\begin{tabular}{@{}c|cccccc@{}}
\toprule
\textbf{Method} & \textbf{Image-Independent} & \textbf{Text-Independent} & \textbf{Transferability} & \textbf{Robustness} & \textbf{Imperceptibility} & \textbf{Efficiency} \\ \midrule
UEs \cite{huang2021unlearnable} & \xmark & \cmark & \xmark & \xmark & \cmark & \xmark \\
RUEs \cite{fu2022robust} & \xmark & \cmark & \xmark & \cmark & \xmark & \xmark \\
TUEs \cite{ren2022transferable} & \xmark & \cmark & \cmark & \xmark & \cmark & \xmark \\
SynPer \cite{yu2022availability} & \cmark & \cmark & \xmark & \xmark & \xmark & \cmark \\
MEM \cite{liu2024multimodal} & \xmark & \xmark & \xmark & \xmark & \cmark & \xmark \\ \midrule
T2UE (Ours) & \cmark & \cmark & \cmark & \cmark & \cmark & \cmark \\ \bottomrule
\end{tabular}
\end{adjustbox}
\end{table*}

\noindent\textbf{Conditional GAN.} Conditional GAN (CGAN) was initially proposed by \citet{2014Conditional}  with the core idea of enabling controlled image generation in GANs, rather than random synthesis. Specifically, CGAN incorporates conditional information into both the generator and discriminator inputs. The generated images must not only appear realistic but also align with the given conditions to be accepted by the discriminator. 
This work has inspired a substantial body of subsequent research. For instance, Radford et al. proposed DCGAN \cite{wu2020dcgan}, which integrated convolutional structures with CGAN to enhance the visual quality of generated images, while StackGAN \cite{zhang2017stackgan} achieved high-resolution text-to-image generation through a two-stage CGAN framework. In cross-modal tasks, models like pix2pix further leveraged paired data to learn image-to-image mapping, where conditional information is provided in the form of input images. Subsequently, the application of conditional generation expanded to diverse scenarios such as video prediction (e.g., VGAN \cite{vondrick2016generating}) and medical image synthesis (e.g., MedGAN \cite{guo2023medgan}). 
The recent study SSGAN \cite{liao2021text} proposed a semantic-spatial aware module to achieve fine-grained text-to-image alignment, significantly enhancing the quality of generated images.  

\section{Text-to-Unlearnable Examples}
\label{sec:method}
We begin by introducing our threat model, followed by an overview of the overall T2UE framework. The detailed components of T2UE are described in subsections \ref{sec:generator} and \ref{sec:objective}.

\begin{figure*}[t]  
    \centering  
    \includegraphics[width=\textwidth]{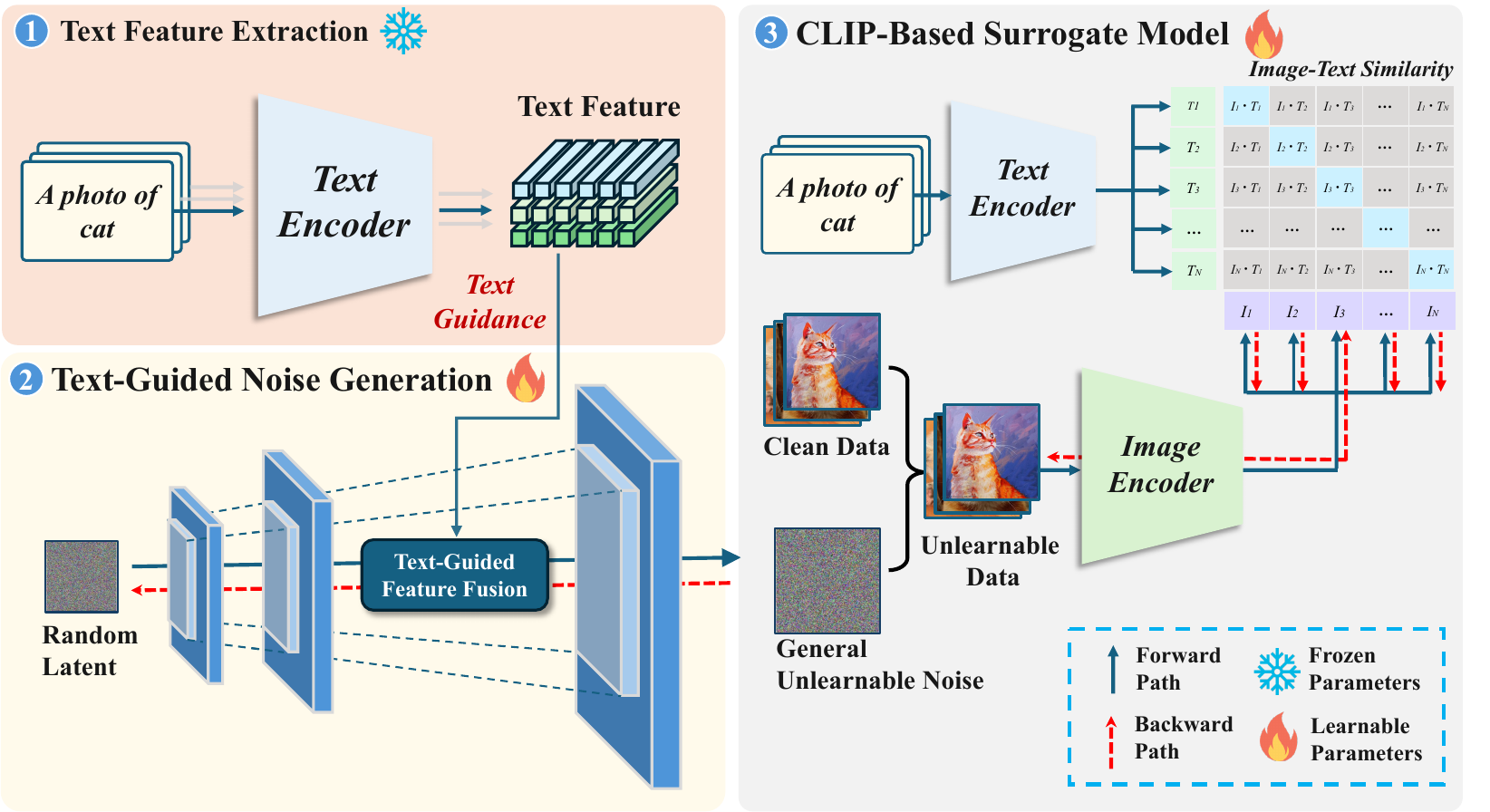}
    \caption{An overview of our proposed T2UE framework, which leverages text description embedding as guidance (stage 1) to guide the generation of general unlearnable noise (stage 2). The noise generator is trained with a surrogate CLIP model to minimize image-text similarity (stage 3).}
    \label{fig:framework_t2ue}
\end{figure*}

\noindent\textbf{Threat Model.}
We consider a scenario involving two primary roles: the \textbf{data protector} (e.g., an individual user or platform) and the \textbf{data hacker} (e.g., an entity scraping data for unauthorized model training). The protector aims to proactively safeguard their multi-modal data (image-text pairs) \textit{before} potential exposure, such as uploading to public platforms or cloud storage. To this end, the protector utilizes a perturbation generation mechanism (potentially leveraging a pre-trained generator model) to transform their original data into Unlearnable Examples (UEs).
Conversely, the hacker gathers these publicly available, yet ostensibly protected, UEs to train their own models. We assume the hacker may employ prevalent training paradigms, including CLIP \cite{radford2021learning} or standard supervised learning for specific downstream tasks. The hacker's objective is to achieve high model performance despite training on the protected dataset.
A critical vulnerability emerges with prior UE generation methods~\citep{liu2024multimodal} on multi-modal data, which  require the protector to access and process the original, sensitive image data to compute the protective perturbation. As highlighted in our Figure~\ref{fig:intro_t2ue}, this workflow necessitates handling or transmitting the very data intended for protection, creating significant  risks such as data theft or leakage during this process.
Therefore, our threat model centers on a protector who seeks a more secure approach: the ability to generate effective UE \textit{without requiring access to the original image data}, ideally leveraging only readily available, less sensitive associated information like textual descriptions. 
The ultimate goal remains unchanged: the generated UEs must effectively disrupt the hacker's model training across both anticipated learning paradigms (CLIP and supervised learning).

\noindent\textbf{Framework Overview.}
The T2UE framework, illustrated in Figure~\ref{fig:framework_t2ue}, is designed to train a generator that maps text semantics to effective unlearnable perturbations. The process involves three key stages. First, \textbf{Text Feature Extraction} utilizes a frozen, pre-trained text encoder (e.g., from CLIP text encoder) to obtain a semantic embedding from the input text. Second, \textbf{Text-Guided Noise Generation} employs the learnable generator network, which takes the text embedding and a random latent vector to synthesize the unlearnable perturbation.

Third, \textbf{CLIP-Based Surrogate Model} guides the optimization of the generator.
In this stage, the unlearnable perturbation is added to the original image to produce a protected image (UE), while the corresponding text descriptions remain unchanged. We adopt the standard InfoNCE contrastive loss and use pre-trained encoders as surrogate models due to their strong generalization capabilities.

\begin{figure}[ht]
  \centering
  \includegraphics[width=0.9\linewidth]{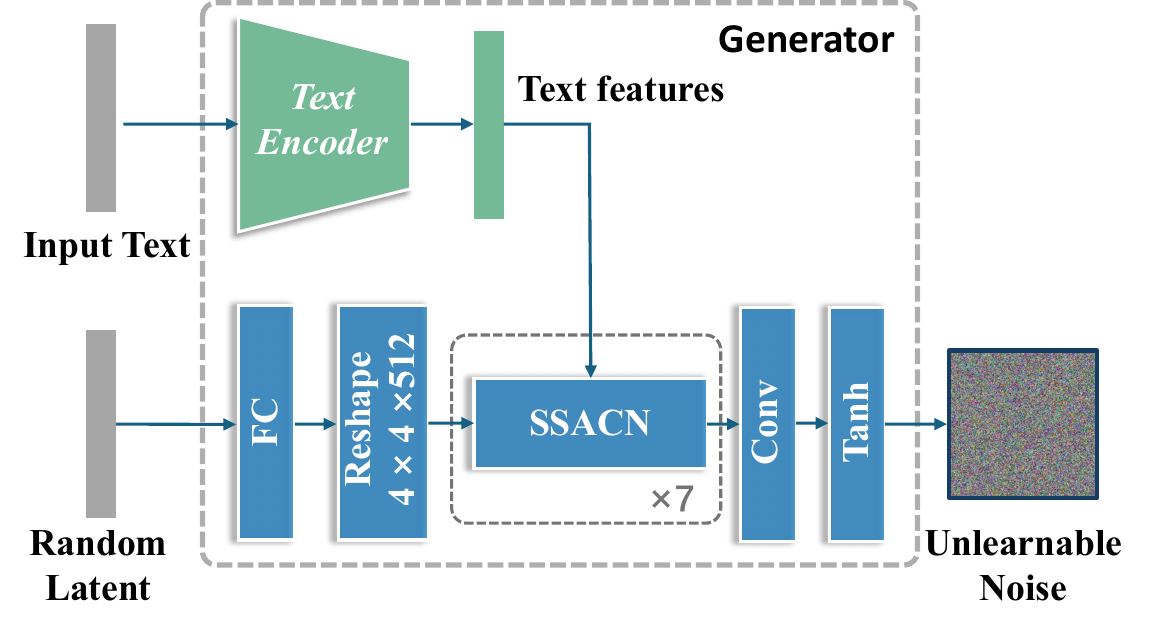}
  \caption{Architecture of T2UE generation model.}  \label{fig:structure_generator}
\end{figure}

\subsection{T2UE Generator Architecture}
\label{sec:generator}

To achieve the mapping from text semantics to unlearnable noise, we instantiate the generator $G$ by adapting the generator architecture from the Semantic-Space-Aware GAN (SSA-GAN)~\cite{liao2022text}, which was originally designed for high-fidelity text-to-image synthesis. We selected it over alternatives like large diffusion models primarily for its balance of effectiveness in conditioning generation on text semantics and its comparatively favourable computational efficiency (reduced parameters, memory footprint, and training time), which is crucial for training the generator within our optimization framework. The generator $G$, parameterized by $\theta_G$, accepts two inputs: the text embedding $\text{emb}_t \in \mathbb{R}^{d_t}$ derived from the frozen text encoder and a random latent vector $z \in \mathbb{R}^{d_z}$ sampled from a standard normal distribution, $z \sim \mathcal{N}(0, I)$. The unlearnable noise is generated by $G$ as follows: 
\begin{gather}
\delta^u = G(\text{emb}_t, z; \theta_G) \quad \text{s.t.} \quad \|\delta^u\|_{\infty} \leq \epsilon.
\label{eq:eq4}
\end{gather}
The core structure of $G$ comprises a sequence of Semantic-Space-Aware Convolutional Network (SSACN) blocks~\cite{liao2022text}. These blocks progressively refine and upsample spatial feature representations, beginning with an initial feature map derived from the latent vector $z$. The key component enabling fine-grained textual control within each SSACN block is Semantic-Space Conditional Batch Normalization (SSCBN). Instead of using standard batch normalization, SSCBN predicts text-specific affine parameters (scale $\gamma$ and shift $\beta$) based on the input text embedding $\text{emb}_t$. These parameters dynamically modulate the normalization of the intermediate feature maps $h$:
\begin{equation}
    \label{eq:sscbn}
    \text{SSCBN}(h, \text{emb}_t) = \gamma(\text{emb}_t) \odot \left( \frac{h - \mu(h)}{\sqrt{\sigma^2(h) + \epsilon_{bn}}} \right) + \beta(\text{emb}_t)
\end{equation}
where $\mu(h)$ and $\sigma^2(h)$ are the mean and variance of $h$ across the batch and spatial dimensions, $\epsilon_{bn}$ is a small constant for numerical stability, $\gamma(\cdot), \beta(\cdot)$ are small neural networks (e.g., linear layers or MLPs) mapping $\text{emb}_t$ to the channel-wise scale and shift parameters, and $\odot$ represents element-wise multiplication. By conditioning the feature statistics on the text embedding at multiple hierarchical levels within the generator, SSCBN allows $G$ to synthesize unlearnable noise patterns $\delta^u$ whose characteristics are semantically correlated with the input text $T$. This semantic correlation is vital for establishing the protective feature intended to mislead the CLIP. Through the sequential application of these SSACN blocks, incorporating residual connections and upsampling operations, the generator synthesizes the final noise perturbation $\delta \in \mathbb{R}^{H \times W \times C}$, typically matching the original image resolution.

\begin{table*}[!t]
\caption{The main results of T2UE's adversarial impact on the CLIP model (with ResNet50 as the vision backbone) in both text-to-image (T2I) and image-to-text (I2T) retrieval tasks. The evaluation covers three benchmark datasets: Flickr8k validation set, Flickr30k, and Tiny-MSCOCO validation set. T2UE demonstrates a significant reduction in the CLIP model's test performance across all datasets. The best protection results are highlighted in \textbf{boldface}.}
\centering
\resizebox{0.9\linewidth}{!}{
\begin{tabular}{@{}c|c|cccc|cccc|cccc@{}}
\toprule
\multirow{3}{*}{Method} & \multirow{3}{*}{\begin{tabular}[c]{@{}c@{}}Zero \\ Contact\end{tabular}} & \multicolumn{4}{c|}{Flick8k} & \multicolumn{4}{c|}{Flick30k} & \multicolumn{4}{c}{Tiny-MSCOCO} \\ \cmidrule(l){3-14} 
 &  & \multicolumn{2}{c}{Image $\rightarrow$ Text} & \multicolumn{2}{c|}{Text $\rightarrow$ Image} & \multicolumn{2}{c}{Image $\rightarrow$ Text} & \multicolumn{2}{c|}{Text $\rightarrow$ Image} & \multicolumn{2}{c}{Image $\rightarrow$ Text} & \multicolumn{2}{c}{Text $\rightarrow$ Image} \\
 &  & Hit@10 & Medr & Hit@10 & Medr & Hit@10 & Medr & Hit@10 & Medr & Hit@10 & Medr & Hit@10 & Medr \\ \midrule
Random Guess & - & 1.2 & 727 & 0.9 & 490 & 1.0 & 711 & 1.0 & 498 & 0.2 & 3419 & 0.2 & 2491 \\
No Protection & -  & 17.7 & 91 & 18.6 & 78 & 49.3 & 11 & 44.1 & 15 & 33.3 & 26 & 15.5 & 247 \\ \midrule
EM & \xmark  & 7.9 & 223 & 8.9 & 154 & 5.5 & 315 & 6.1 & 190 & 3.7 & 402 & 3.9 & 344 \\
TAP & \xmark  & 14.3 & 120 & 12.8 & 92 & 39.7 & 19 & 37.2 & 23 & 7.4 & 194 & 6.0 & 289 \\
MEM-3 & \xmark & 15.9 & 92 & 17.5 & 78 & 47.6 & 12 & 43.5 & 16 & 28.5 & 31 & 15.5 & 236 \\
MEM-5 & \xmark & 17.6 & 92 & 17.7 & 77 & 47.1 & 13 & 43.3 & 16 & 29.2 & 30 & 15.4 & 246 \\ \midrule
\rowcolor{gray!10} T2UE (Ours)
& \cmark & \textbf{11.6\textcolor{red}{\small{(6.1↓)}}} & \textbf{155\textcolor{red}{\small{(68↓)}}} & \textbf{8.02\textcolor{red}{\small{(10.56↓)}}} & \textbf{194\textcolor{red}{\small{(116↓)}}}  
& \textbf{18.2\textcolor{red}{\small{(31.1↓)}}} & \textbf{78\textcolor{red}{\small{(67↓)}}} & \textbf{15.62\textcolor{red}{\small{(28.48↓)}}} & \textbf{106\textcolor{red}{\small{(91↓)}}}  
& \textbf{7.4\textcolor{red}{\small{(25.9↓)}}} & \textbf{220\textcolor{red}{\small{(194↓)}}} & \textbf{5.14\textcolor{red}{\small{(10.38↓)}}} & \textbf{331\textcolor{red}{\small{(84↓)}}}
\\ \bottomrule
\end{tabular}}
\label{tab:clip_pretrained}
\end{table*}

\begin{table}[!t]
  \caption{Performance comparison of state-of-the-art methods in supervised learning. Experiments are conducted on CIFAR-10 and CIFAR-100 using ResNet-18 as the target model. \textbf{T2UE (ViT/B-16)} denotes that the generator aligns with CLIP’s ViT-B/16 model, while \textbf{T2UE (ViT/B-32)} indicates alignment with CLIP’s ViT-B/32.
  T2UE achieves competitive performance with the best baseline method while maintain zero-contact to the original images.}
  \label{tab:supervised_result}
  \centering
  \resizebox{1.0\linewidth}{!}{
  \begin{tabular}{@{}cccccc@{}}
\toprule
\multirow{2}{*}{Method} & \multirow{2}{*}{\begin{tabular}[c]{@{}c@{}}Zero \\ Contact\end{tabular}} & \multicolumn{2}{c}{CIFAR-10} & \multicolumn{2}{c}{CIFAR-100} \\ \cmidrule(l){3-6} 
 &  & Class-wise & Sample-wise & Class-wise & Sample-wise \\ \midrule
No Protection & \cmark & 90.34 & 90.34 & 65.98 & 65.98 \\ \midrule
EM & \xmark & 11.57 & 15.37 & \textbf{1.07} & 3.21 \\
Emax & \xmark & - & 45.39 & - & 47.69 \\
TUE & \xmark & - & \textbf{10.06} & - & \textbf{1.05} \\
UC & \xmark & 79.41 & - & 41.38 & - \\
AR & \cmark & 10.41 & - & 5.11 & - \\
SynPer & \cmark  & - & 17.00 & - & 1.89 \\
TAP & \xmark & - & 10.78 & - & 60.60 \\
CP & \xmark & 10.12 & 83.15 & 1.31 & 49.91 \\ \midrule
\textbf{T2UE(ViT/B-16)} & \cmark & \textbf{10.02\textcolor{red}{\small{(80.32↓)}}} & 13.76\textcolor{red}{\small{(76.58↓)}} & 1.08\textcolor{red}{\small{(64.81↓)}} & 1.82\textcolor{red}{\small{(64.16↓)}} \\
\textbf{T2UE(ViT/B-32)} & \cmark & 12.17\textcolor{red}{\small{(78.17↓)}} & 13.89\textcolor{red}{\small{(76.45↓)}} & 1.42\textcolor{red}{\small{(64.56↓)}} & 1.85\textcolor{red}{\small{(64.13↓)}} \\ \bottomrule
\end{tabular}
  }
\end{table}

\begin{table*}[!t]
  \caption{\small Transferability evaluation of UE across different model Architectures. The noise for all baseline methods is generated using the ResNet-50 model and evaluated on various network architectures (ResNet-18/50\cite{he2016deep}, VGG-11/13\cite{simonyan2014very}, GoogLeNet\cite{szegedy2015going}, DenseNet-121\cite{huang2017densely}) on the CIFAR-10 and CIFAR-100 datasets. T2UE demonstrates superior transferability across different model backbones.}
  \label{tab:supervised_result_across_model}
  \centering
  \begin{adjustbox}{width=0.9\linewidth}
\begin{tabular}{c|cccc|cccc}
\toprule
\multirow{2}{*}{Method} & \multicolumn{2}{c}{CIFAR-10} & \multicolumn{2}{c|}{CIFAR-100} & \multicolumn{2}{c}{CIFAR-10} & \multicolumn{2}{c}{CIFAR-100} \\ \cmidrule{2-9} 
 & Class-wise & Sample-wise & Class-wise & Sample-wise & Class-wise & Sample-wise & Class-wise & Sample-wise \\ \midrule
\rowcolor{gray!10} Target Model & \multicolumn{4}{c|}{Resnet18} & \multicolumn{4}{c}{Resnet50} \\ \midrule
No Protection & 90.34 & 90.34 & 65.98 & 65.98 & 85.56 & 85.56 & 47.44 & 47.44 \\ \midrule
EM & 11.57 & 15.37 & \textbf{1.07} & 3.21 & 12.81 & 11.40 & \textbf{0.88} & 3.10 \\
Emax & - & 45.39 & - & 47.69 & - & 38.67 & - & 41.76 \\
TUE & - & \textbf{10.06} & - & \textbf{1.05} & - & \textbf{10.18} & - & \textbf{0.85} \\
UC & 79.41 & - & 41.38 & - & 79.18 & - & 37.28 &  \\
AR & 10.41 & - & 5.11 & - & 10.92 & - & 2.18 &  \\
SynPer & - & 17.00 & - & 1.89 & - & 13.06 & - & 1.79 \\
TAP & - & 10.78 & - & 60.6 & - & 22.80 & - & 43.55 \\
CP & 10.12 & 83.15 & 1.31 & 49.91 & 10.27 & 84.42 & 1.05 & 42.61 \\ \midrule
\textbf{T2UE(ViT/B-16)} & \textbf{10.02\textcolor{red}{\small{(80.32↓)}}} & 13.76\textcolor{red}{\small{(76.58↓)}} & 1.08\textcolor{red}{\small{(64.81↓)}} & 1.82\textcolor{red}{\small{(64.16↓)}} & \textbf{10.04\textcolor{red}{\small{(75.52↓)}}} & 10.97\textcolor{red}{\small{(75.59↓)}} & 1.07\textcolor{red}{\small{(46.37↓)}} & 1.36\textcolor{red}{\small{(46.08↓)}} \\
\textbf{T2UE(ViT/B-32)} & 12.17\textcolor{red}{\small{(78.17↓)}} & 13.89\textcolor{red}{\small{(76.45↓)}} & 1.42\textcolor{red}{\small{(64.56↓)}} & 1.85\textcolor{red}{\small{(64.13↓)}} & 10.41\textcolor{red}{\small{(75.15↓)}} & 10.41\textcolor{red}{\small{(75.15↓)}} & 1.18\textcolor{red}{\small{(46.26↓)}} & 1.21\textcolor{red}{\small{(46.23↓)}} \\ \midrule
\rowcolor{gray!10} {Target Model} & \multicolumn{4}{c|}{VGG11} & \multicolumn{4}{c}{VGG13} \\ \midrule
No Protection & 84.65 & 84.65 & 57.78 & 57.78 & 86.30 & 86.30 & 61.50 & 61.50 \\ \midrule
EM & 15.31 & \textbf{12.90} & \textbf{1.07} & 3.63 & 13.40 & \textbf{12.03} & \textbf{1.00} & 3.60 \\
Emax & - & 80.36 & - & 57.39 & - & 80.58 & - & 58.17 \\
TUE & - & 13.53 & - & \textbf{1.76} & - & 12.06 & - & \textbf{1.56} \\
UC & 82.31 & - & 56.94 & - & 85.43 & - & 56.83 & - \\
AR & 29.55 & - & 50.18 & - & \textbf{10.32} & - & 28.46 & - \\
SynPer & - & 18.94 & - & 4.19 & - & 15.68 & - & 3.43 \\
TAP & - & 50.46 & - & 57.39 & - & 40.55 & - & 43.15 \\
CP & 11.13 & 83.31 & 2.45 & 57.50 & 11.11 & 86.29 & 1.84 & 58.69 \\ \midrule
\textbf{T2UE(ViT/B-16)} & \textbf{10.96\textcolor{red}{\small{(73.69↓)}}} & 28.43\textcolor{red}{\small{(56.22↓)}} & 3.58\textcolor{red}{\small{(54.20↓)}} & 9.55\textcolor{red}{\small{(48.23↓)}} & 11.93\textcolor{red}{\small{(74.37↓)}} & 19.73\textcolor{red}{\small{(66.57↓)}} & 1.21\textcolor{red}{\small{(46.37↓)}} & 6.32\textcolor{red}{\small{(55.18↓)}} \\
\textbf{T2UE(ViT/B-32)} & 15.65\textcolor{red}{\small{(69.06↓)}} & 24.58\textcolor{red}{\small{(59.98↓)}} & 2.89\textcolor{red}{\small{(54.89↓)}} & 15.08\textcolor{red}{\small{(42.70↓)}} & 10.44\textcolor{red}{\small{(75.86↓)}} & 18.42\textcolor{red}{\small{(67.88↓)}} & 1.24\textcolor{red}{\small{(60.26↓)}} & 16.13\textcolor{red}{\small{(45.37↓)}} \\ \midrule
\rowcolor{gray!10} {Target Model} & \multicolumn{4}{c|}{GoogLeNet} & \multicolumn{4}{c}{DenseNet121} \\ \midrule
No Protection & 89.89 & 89.89 & 69.74 & 69.74 & 87.81 & 87.81 & 70.29 & 70.29 \\ \midrule
EM & 11.93 & 17.75 & \textbf{1.00} & 3.40 & 12.86 & 13.86 & 1.14 & 3.42 \\
Emax & - & 25.61 & - & 67.88 & - & 41.85 & - & 64.33 \\
TUE & - & 11.94 & - & \textbf{1.00} & - & \textbf{12.07} & - & \textbf{1.05} \\
UC & 80.29 & - & 40.38 & - & 85.99 & - & 66.52 & - \\
AR & 11.76 & - & 4.91 & - & 12.27 & - & 4.68 & - \\
SynPer & - & 17.81 & - & 10.00 & - & 13.42 & - & 2.12 \\
TAP & - & 23.01 & - & 65.86 & - & 19.97 & - & 66.66 \\
CP & 11.67 & 55.85 & 1.24 & 67.99 & 10.51 & 86.69 & 1.69 & 67.87 \\ \midrule
\textbf{T2UE(ViT/B-16)} & \textbf{9.44\textcolor{red}{\small{(80.45↓)}}} & \textbf{11.44\textcolor{red}{\small{(78.45↓)}}} & 1.06\textcolor{red}{\small{(68.68↓)}} & 1.94\textcolor{red}{\small{(67.8↓)}} & \textbf{10.02\textcolor{red}{\small{(77.79↓)}}} & 13.21\textcolor{red}{\small{(74.6↓)}} & \textbf{1.11\textcolor{red}{\small{(69.18↓)}}} & 3.06\textcolor{red}{\small{(67.23↓)}} \\
\textbf{T2UE(ViT/B-32)} & 10.35\textcolor{red}{\small{(79.54↓)}} & 11.29\textcolor{red}{\small{(78.6↓)}} & 1.37\textcolor{red}{\small{(68.37↓)}} & 2.82\textcolor{red}{\small{(66.92↓)}} & 12.12\textcolor{red}{\small{(75.69↓)}} & 17.08\textcolor{red}{\small{(70.73↓)}} & 1.57\textcolor{red}{\small{(68.72↓)}} & 3.02\textcolor{red}{\small{(67.27↓)}} \\ \bottomrule
\end{tabular}
\end{adjustbox}
\end{table*}

\begin{table*}[!t]
  \caption{\small Mixed Poisoning Robustness Evaluation on CIFAR-10/100 with ResNet-50 (0\% Clean to 100\% Poisoned).}
  \label{tab:supervised_different_percentage_exp}
  \centering
  \resizebox{0.9\linewidth}{!}{
  \begin{tabular}{c|cccc|cccc}
    \toprule
    {Dataset} & \multicolumn{2}{c}{CIFAR-10} & \multicolumn{2}{c}{CIFAR-100} & \multicolumn{2}{c}{CIFAR-10} & \multicolumn{2}{c}{CIFAR-100} \\
    \midrule[0.75pt]
    {T2UE Version}
    & \multicolumn{2}{c}{T2UE(ViT/B-16)} & \multicolumn{2}{c}{T2UE(ViT/B-32)}  & \multicolumn{2}{c}{T2UE(ViT/B-16)} & \multicolumn{2}{c}{T2UE(ViT/B-32)}       \\
    \midrule
    {Method}  & Class-wise & Sample-wise & Class-wise & Sample-wise & Class-wise & Sample-wise & Class-wise & Sample-wise \\
    \midrule
    No Protection & 90.34 & 90.34 & 65.98 & 65.98 
               & 90.34 & 90.34 & 65.98 & 65.98
    \\ 
    20\%   & 79.46 & 82.65 & 52.68 & 49.90 
           & 81.22 & 82.37 &  47.3 & 46.87
    \\
    40\%   & 78.01 & 81.00 & 45.33 & 46.15 
           & 79.11 & 78.64 & 43.11 & 41.27
    \\
    60\%   & 74.46 & 77.88 & 34.76 & 41.16                   & 74.24 & 74.25 & 35.56 & 33.63
    \\
    80\%   & 68.44 & 70.37 & 26.27 & 28.82                   & 64.57 & 64.65 & 24.22 & 26.72
    \\
   100\%  & \textbf{10.02} & \textbf{13.76} & \textbf{1.08} & \textbf{1.82} & \textbf{12.17} & \textbf{13.89} & \textbf{1.42} & \textbf{1.41}
    \\
    
    \bottomrule
  \end{tabular}
  }
\end{table*}

\subsection{T2UE Generator Training}
\label{sec:objective}

The objective of T2UE is to train the text-guided generator $G$, that it produces image perturbations $\delta^u$ that render data attacker's training ineffective. We achieve this by optimizing $G$ using a frozen pre-trained CLIP model $(f_I, f_T)$, as a surrogate to simulate potential data exploitation, as illustrated in Figure \ref{fig:framework_t2ue}. Formally, given a training data distribution $\mathcal{D}$ of image-text pairs $(I, T)$, the objective of standard multi-modal contrastive learning is to train image ($f_I$) and text ($f_T$) encoders to align representations of corresponding pairs while distinguishing non-corresponding pairs within a shared embedding space. This alignment is commonly achieved by minimizing a contrastive loss over batches of $N$ pairs, such as the symmetric InfoNCE loss:
\begin{equation}
\label{eq:standard_clip_loss}
\min_{\theta_I, \theta_T} \mathbb{E}_{\{(I_i, T_i)\}_{i=1}^N \sim \mathcal{D}} \left[ \mathcal{L}_{\text{InfoNCE}}\left( \{f_I(I_i; \theta_I)\}_{i=1}^N, \{f_T(T_i; \theta_T)\}_{i=1}^N \right) \right].
\end{equation}
Our T2UEs aims to establish a strong, protective features between the input text $T$ and the generated $\delta^u$, thereby disrupting the model's ability to learn the true relationship between original images $I$ and texts $T$. Thus, the optimization objective is to find the optimal generator parameters $\theta_G^*$ by minimizing the expected alignment loss computed by the surrogate model on protected data:
\begin{equation}
\label{eq:objective}
\theta_G^* = \arg\min_{\theta_G} \mathbb{E}_{(I,T) \sim \mathcal{D}} \left[ \mathcal{L}_{\text{InfoNCE}}\left( \{f_I(I + \delta^u)\}, \{f_T(T)\} \right) \right].
\end{equation}
Minimizing this objective (Eq.~\ref{eq:objective}) compels the generator $G$ to synthesize perturbations $\delta^u$ that, when added to any image $I$, cause the resulting image feature $f_I(I+\delta')$ to align strongly with the feature $f_T(T)$ corresponding to the conditioning text $T$. Consequently, models subsequently trained on datasets $\{(I+\delta', T)\}$ learn this easily predictable feature instead of the meaningful image-text relationships, thus achieving the unlearnable effect.

\noindent\textbf{Zero-Contact Data Protection.}
Once trained, the generator produces unlearnable noise $\delta^u$ based solely on a text prompt $T$ and a random vector $z$, without requiring access to the target image. This noise can then be applied to any image for downstream tasks. For classification, specific noise patterns can be generated as needed—for example, class-wise patterns using fixed templates and $z$, or sample-wise patterns using varied templates and $z$—demonstrating the flexibility of the text-guided approach.

\section{Experiment}

\noindent\textbf{Datasets and Target Models.}
Our T2UE generator was trained on the MSCOCO dataset~\cite{lin2014microsoft}. We evaluated its effectiveness in disrupting CLIP-like models (using ResNet50~\cite{he2016deep} or ViT-B/32~\cite{dosovitskiy2020vit} image backbones) from scratch on protected versions of Flickr8k~\cite{young2014image}, Flickr30k~\cite{young2014image}, and Tiny-MSCOCO~\cite{chen2015microsoft}. Transferability to supervised learning was assessed by training standard classification backbones (ResNet18/50/101~\cite{he2016deep}, ViT-B/16/32~\cite{dosovitskiy2020vit}) on protected CIFAR-10/-100~\cite{krizhevsky2009learning} and STL-10~\cite{coates2011analysis}.

\noindent\textbf{Implementation Details.}
The T2UE generator leverages a CLIP ViT-B/32~\cite{radford2021learning} alignment model and was trained for 500 epochs (ViT-B/16 variant: 300 epochs). For the generator training stage, following the training setup of the generator model in \cite{liao2021text}, we choose 
a batch size of 128 and we employ an Adam optimizer with initial learning rate to 0.0001 and decoupled weight decay regularization, while the learning
rate decays using a cosine scheduler. We train these models from
scratch on 1 A100 GPU for 32 epochs. Generated unlearnable noise $\delta^u$ were constrained with $||\delta||_{\infty} \le 8/255$. Target supervised models were trained for 100 epochs. Following the parameters in \cite{huang2021unlearnable}, we use SGD with a learning rate of 0.1, momentum (0.9), and weight decay ($5.0 \times 10 ^ {-4}$), along with cosine annealing scheduling of the learning rate 
For fair comparison, target model architectures and training setups were kept consistent across compared methods within each experiment.

\textbf{Baselines.}
In the CLIP setting, we compare T2UE with three applicable baselines: \textbf{EM}\cite{huang2021unlearnable}, \textbf{TAP}\cite{fowl2021adversarial}, and \textbf{MEM}~\cite{liu2024multimodal}.
 As our setup simulates user-side application, we adapt MEM by using only its image noise (MEM-3/MEM-5 denote variants omitting 3/5-token triggers). In the supervised setting, we compare against established UE methods including 
\textbf{EM}~\cite{zhang2023unlearnable}, \textbf{Emax}~\cite{koh2017understanding}, 
\textbf{AR}~\cite{sandoval2022autoregressive}, \textbf{TUE}~\cite{ren2022transferable}, 
\textbf{UC}~\cite{zhang2023unlearnable}, \textbf{SynPer}~\cite{yu2022availability}, 
\textbf{TAP}~\cite{fowl2021adversarial}, \textbf{CP}~\cite{he2023indiscriminate}, 
using ResNet18 as the surrogate model where required.

\noindent\textbf{Evaluation Metrics.}
For MCL, we measure zero-shot retrieval performance degradation using Hit@10 (Recall@10) and Median Rank (MedR) for I2T and T2I tasks~\cite{yang2024robust, yang2023data}; lower Hit@10 and higher MedR indicate better protection. For supervised learning, we report Top-1 accuracy on clean test sets; lower accuracy indicates better protection.

Appendix~\ref{appendix:generator impact} presents an ablation study on the generator, while Appendix~\ref{appendix:efficiency} offers a computational cost analysis, showing that T2UE achieves significant efficiency advantages.

\begin{table*}[!t]
\caption{\small{The test accuracy (\%) of ResNet50 trained using different defense methods on unlearnable CIFAR-10 and CIFAR-100 crafted by T2UE.}}
\label{tab:different_denfense}
\centering
\resizebox{\textwidth}{!}{
\begin{tabular}{c|cccc|cccc}
\toprule
Dataset     & \multicolumn{4}{c|}{CIFAR-10} & \multicolumn{4}{c}{CIFAR-100} \\ \midrule
Defense      & None  & CutOut   & MixUp   & AutoAug    & None  & CutOut   & MixUp   & AutoAug    \\ \midrule
No Protection  & 83.91 & 83.14 & 86.17 & 83.53 &  65.98 & 56.59 & 49.88 & 53.78 \\ 
T2UE(ViT/B-16)-C & \textbf{10.02} &  9.99 & \textbf{11.44} & \textbf{10.02} &  \textbf{1.08} & \textbf{1.62} & 1.51 &  \textbf{1.03} \\ 
T2UE(ViT/B-16)-S & 13.76 & 10.97 & 10.33 & 14.35 &  1.82 &  1.64 & 2.06 &  1.81 \\ 
T2UE(ViT/B-32)-C & 12.17 & 11.21 & \textbf{9.98} & 11.97 &  1.42 &  1.15 & \textbf{0.98} &  1.57 \\ 
T2UE(ViT/B-32)-S & 13.89 & \textbf{12.30} & 10.56 & 11.03 &  1.41 &  1.77 & 1.46 &  1.64 \\ 
\bottomrule
\end{tabular}
}
\end{table*}

\begin{figure*}[!t]
	\centering
	\begin{subfigure}{0.24\linewidth}
		\includegraphics[width=\textwidth]{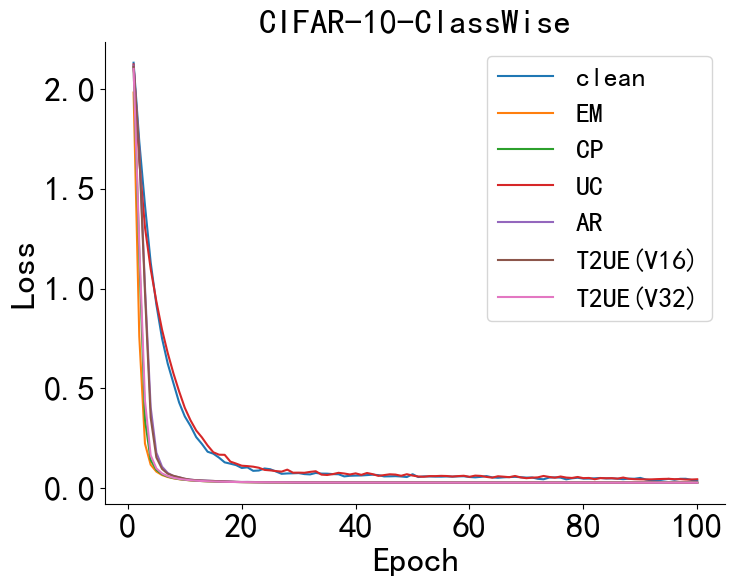}
		\caption{\tiny{Class-wise $\Delta_c$ Loss}}
		\label{fig:supervised_train_loss_CIFAR-10_classwise_subfig}
	\end{subfigure}
	\begin{subfigure}{0.24\linewidth}
		\includegraphics[width=\textwidth]{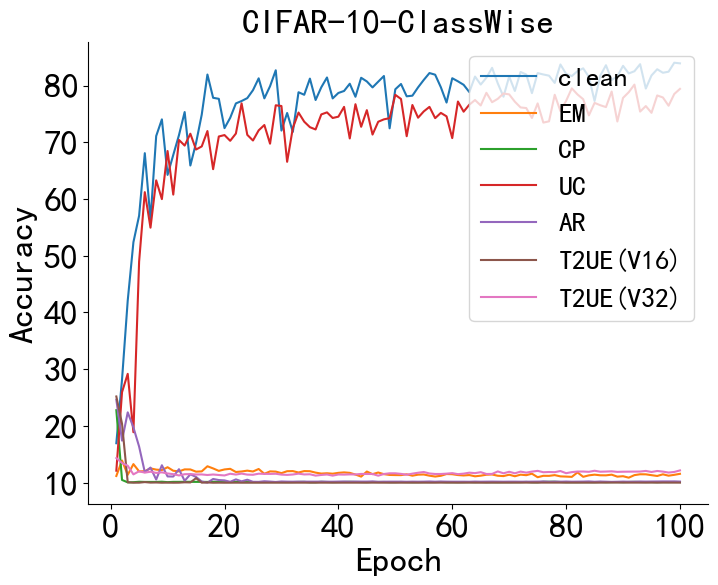}
		\caption{\tiny{Class-wise $\Delta_c$ Accuracy}}
		\label{fig:supervised_test_acc_train_loss_CIFAR-10_classwise_subfig}
	\end{subfigure}
	\begin{subfigure}{0.24\linewidth}
		\includegraphics[width=\textwidth]{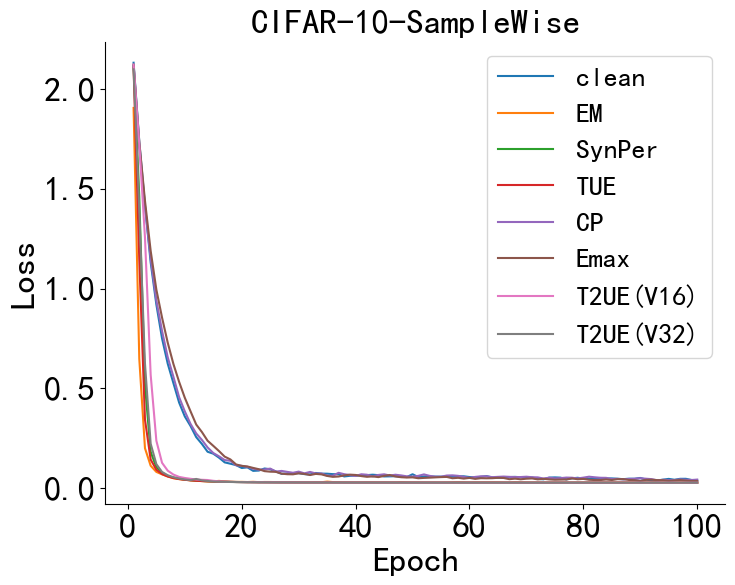}
		\caption{\tiny{Sample-wise $\Delta_s$ Loss}}
		\label{fig:supervised_train_loss_CIFAR-10_samplewise_subfig}
	\end{subfigure}
	\begin{subfigure}{0.24\linewidth}
		\includegraphics[width=\textwidth]{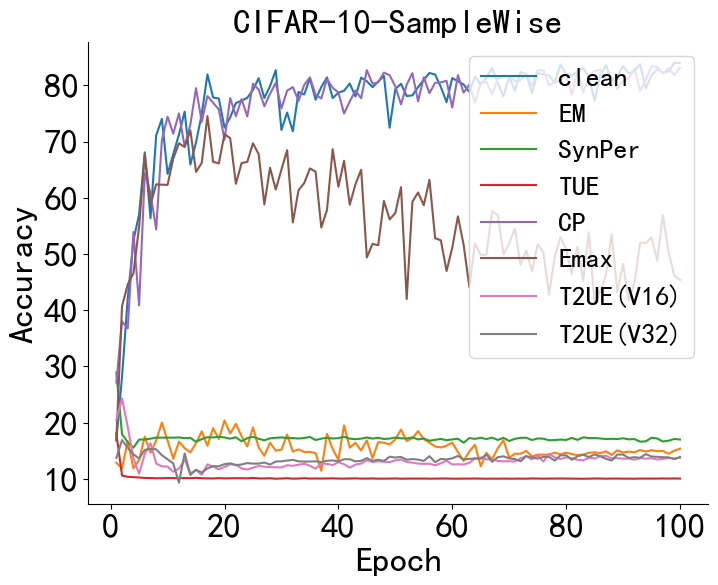}
		\caption{\tiny{Sample-wise $\Delta_s$ Accuracy}}
		\label{fig:supervised_test_acc_CIFAR-10_samplewise_subfig}
	\end{subfigure}
	\caption{(a-b): For class-wise noise, learning curves of ResNet50 on CIFAR-10 dataset with different types of training data: 1) Loss curve, 2) Test Accuracy Curve. (c-d): For sample-wise noise, learning curves of ResNet50 on CIFAR-10 dataset.}
	\label{fig:training_curve_cifar10}
\end{figure*}

\subsection{Main Results}
\label{sec:exp_main}

\noindent\textbf{Effectiveness on CLIP.} We evaluate T2UE's effectiveness against CLIP using zero-shot retrieval metrics (Hit@10/Medr)
on Flickr8k, Flickr30k, and Tiny-MSCOCO after pre-training a ResNet50-based CLIP model (Table \ref{tab:clip_pretrained}). Critically, the MEM method~\cite{liu2024multimodal}, when applied using only its image noise (MEM-3/MEM-5, simulating a realistic user scenario without text trigger modification), shows negligible protection effect compared to the `no protection' baseline, highlighting its dependence on fragile text triggers. TAP~\cite{fowl2021adversarial} provides some protection but is inconsistent, notably degrading on the larger Flickr30k dataset. While the image-dependent EM~\cite{huang2021unlearnable} method severely degrades performance, it requires image access, violating our core privacy constraint. In contrast, T2UE consistently delivers the strongest protection among the image-agnostic methods across all datasets. For instance, on Flickr30k, T2UE drastically reduces I2T Hit@10 to 18.2, far surpassing TAP (39.7) and MEM-5 (47.1). This demonstrates that T2UE achieves substantial disruption of CLIP training entirely without image access, effectively fulfilling the \textbf{"zero-contact"} requirement while providing significantly better protection than comparable baselines.

\noindent\textbf{Effectiveness of Transferring to Supervised Learning.} We evaluate the transferability of T2UE’s protective noise to standard supervised classification tasks. Specifically, we directly apply the T2UE Generator to supervised learning scenarios on CIFAR-10 and CIFAR-100. Table~\ref{tab:supervised_result} presents the Top-1 accuracy achieved by a ResNet-18 model trained for 100 epochs on data perturbed by T2UE and various baselines. 
T2UE(ViT/B-16) and T2UE(ViT/B-32) refer to variants generated using CLIP ViT-B/16 and ViT-B/32 models, respectively. The results clearly demonstrate the effectiveness of T2UE in the supervised setting. On CIFAR-10, T2UE(ViT/B-16) reduces the model’s accuracy to 10.02\%, matching the performance of top class-wise methods like CP (10.12\%) and leading sample-wise methods such as TUE (10.06\%). On CIFAR-100, T2UE(ViT/B-16) achieves an accuracy of 1.08\%, closely aligning with the strongest class-wise baseline EM (1.07\%) and the sample-wise baseline TUE (1.05\%). 
Notably, T2UE attains this highly competitive performance \textit{without any access} to CIFAR images during noise generation, relying solely on text guidance. More importantly, this result underscores the strong transferability of T2UE’s unlearnable noise, which remains effective even outside its targeted training framework. The transferability to different tasks was a critical bottleneck in prior work \cite{huang2021unlearnable}.

\noindent\textbf{Effectiveness of Transferring to Different Architectures.}
Table~\ref{tab:supervised_result_across_model} presents the results of transfer attack experiments using different models. As shown, the T2UE consistently maintains the unlearnability effect across various models. Notably, sample-level noise exhibits higher model dependency than class-level noise. For sample-level noise, the classification accuracy on unlearnable data remains relatively stable regardless of the training model. In contrast, class-level noise tends to show improved model accuracy when transferred to different architectures. These results demonstrate that T2UE can simultaneously achieve cross-task and cross-model transferability of unlearnable noise, all while maintaining complete “zero-contact” with the original data.

\subsection{Robustness Evaluations}

In this subsection, we perform a standard evaluation of UE by examining: (1) the impact of varying proportions of unlearnable data in the training set, and (2) its robustness to data augmentation.

\noindent\textbf{Different proportion of unlearnable data in the training set.}
Table~\ref{tab:supervised_different_percentage_exp} presents the poisoning effects of T2UE noise on ResNet-18. The experiments were conducted on CIFAR-10/100 with poisoning ratios ranging from 20\% to 100\%. The results demonstrate that higher poisoning ratios lead to more pronounced effects. For instance, the accuracy on CIFAR-10 drops from 79.46\% at 20\% poisoning to 10.02\% at 100\% poisoning, indicating that even partial poisoning can effectively degrade model performance.

\noindent\textbf{Robustness against data augmentation.} Following prior works~\cite{huang2021unlearnable,zhang2023unlearnable}, we conduct a robustness evaluation using commonly adopted data augmentation techniques, including CutOut~\cite{devries2017cutout}, MixUp~\cite{zhang2017mixup}, and AutoAugment (AutoAug)\cite{8953317}. Table~\ref{tab:different_denfense} presents the performance of T2UE when these augmentations are applied during the training process.
T2UE consistently preserves the unlearnability effect across various data augmentation scenarios, demonstrating its robustness in bypassing a range of augmentation strategies and preventing the model from learning meaningful data representations.

\noindent\textbf{Training curves under supervised learning.} 
Figure~\ref{fig:training_curve_cifar10} illustrates the training loss and test accuracy over epochs for the ResNet18 model. The “clean” curve represents training on unaltered data, while the others correspond to various poisoned datasets. Except for UC, all methods converge to a test accuracy close to random guessing (10\%) by the 100th epoch. Although UC achieves higher accuracy, its training loss closely mirrors that of clean training, indicating a failure to induce true unlearnability.

In contrast, our proposed T2UE method, along with other unlearnable baselines, demonstrates rapid convergence in training loss while maintaining consistently low test accuracy—evidence of effective poisoning. Notably, T2UE sustains low inference accuracy even in the early stages of training, outperforming methods like AR, which are susceptible to early stopping strategies. Furthermore, T2UE preserves strong unlearnability at the sample level.

\noindent\textbf{Varying text descriptions.}
In practice, different users may focus on varying aspects of the same image, the textual descriptions they construct can differ significantly. To better align with real-world  scenarios, we explore whether our T2UE can still be effective under such settings.
Specifically, this study utilizes the GIT model \cite{wang2022git} released by Microsoft to generate diverse textual descriptions for each image in the dataset, thereby expanding and enriching the original training data. To further enhance the diversity of generated descriptions, a series of image augmentation techniques—including random cropping, horizontal flipping, random rotation, and other common methods—are applied to each image, improving the model's adaptability to various scenarios.  
The experimental results are presented in Table~\ref{tab:image2text_effect}. They show that T2UE is robust to variations in text descriptions, consistently providing effective protection under these conditions.

\noindent\textbf{Additional Ablation Studies.} In Appendix~\ref{appendix:ablation}, we provide further ablation studies on the effect of generator training duration, along with an efficiency analysis. The results show that (1) the generator progressively improves and achieves strong performance given sufficient initial training, and (2) our T2UE is notably more efficient than existing approaches.

\begin{table}[!hbt]
\centering
\caption{Evaluation results on varying different text descriptions. Lower Hit / higher MedR indicates better protection. }
\label{tab:image2text_effect}
\centering
\begin{adjustbox}{width=1.0\linewidth}
\begin{tabular}{@{}cccccccccc@{}}
\toprule
 &  & \multicolumn{4}{c}{Image $\rightarrow$ Text} & \multicolumn{4}{c}{Text $\rightarrow$ Image} \\ \midrule
Dataset & Method & Hit@1 & Hit@5 & Hit@10 & Medr & Hit@1 & Hit@5 & Hit@10 & Medr \\ \midrule
\multirow{2}{*}{Flickr8K} & Clean & 1.0 & 3.6 & 5.4 & 226 & 0.54 & 2.7 & 5.2 & 258 \\
 & T2UE & 0.3 & 2.9 & 5.6 & 249 & 0.48 & 2.32 & 4.92 & 267 \\ \midrule
\multirow{2}{*}{Flickr30K} & Clean & 1.2 & 5.4 & 8.1 & 224 & 0.82 & 3.24 & 5.7 & 265 \\
 & T2UE & 0.6 & 2.3 & 3.0 & 465 & 0.76 & 3.08 & 4.94 & 291 \\ \midrule
\multirow{2}{*}{Tiny-MSCOCO} & Clean & 0.7 & 3.1 & 4.9 & 301 & 0.46 & 2.00 & 3.46 & 373 \\
 & T2UE & 0.6 & 1.9 & 2.8 & 427 & 0.36 & 1.66 & 2.92 & 392 \\ \bottomrule
\end{tabular}

\end{adjustbox}
\end{table}

\subsection{Visualization}
\label{appendix:visualization}

\begin{figure}[h]  
    \centering  
    \includegraphics[width=1.0\linewidth]{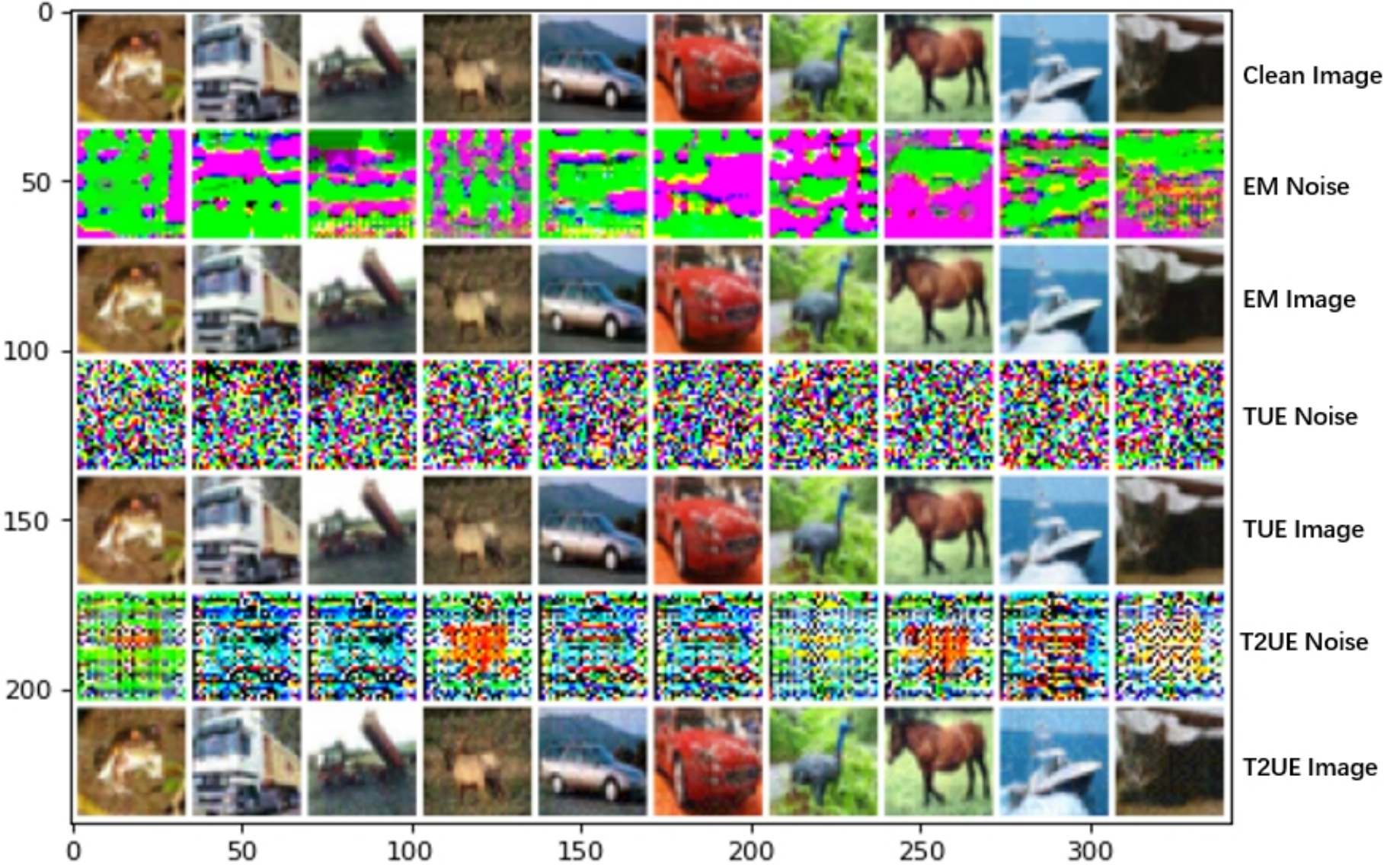}
    \caption{Visualization of different UEs and their corresponding protective noise generated on CIFAR-10. The noise patterns have been normalized for visualization purposes.}  \label{fig:visualization_all_method}
\end{figure}

In Figure~\ref{fig:visualization_all_method}, we present visualizations of different unlearnable example (UE) methods, including EM~\cite{huang2021unlearnable}, TUE~\cite{ren2022transferable}, and our proposed T2UE. Notably, T2UE produces distinctive noise patterns that are visually different from those generated by other methods.


\section{Limitation and Conclusion}
In this paper, we introduced Text-to-Unlearnable-Example (T2UE), a novel image-independent framework that generates strong protective perturbations without requiring access to the original image, thereby achieving \textit{zero-contact} data protection. By learning a mapping from text semantics to the perturbation space, T2UE effectively decouples noise generation from sensitive visual content.
We demonstrated that T2UE can effectively disrupt CLIP, outperforming existing multimodal unlearnable example baselines. Its protective effect also transfers remarkably well to supervised classification tasks, achieving performance comparable to strong image-dependent methods—all without accessing the images themselves. In addition, T2UE provides substantial computational efficiency gains.
By addressing the core limitation of image dependency, T2UE presents a practical, secure, and effective solution for user-centric protection against unauthorized multimodal learning, paving the way for more trustworthy multimodal data protection.

\noindent While T2UE offers a promising image-independent approach to generating unlearnable examples, several limitations remain. Its effectiveness can be sensitive to the quality and consistency of input text descriptions—poorly constructed or ambiguous prompts may weaken the protective effect.  Additionally, while T2UE achieves performance comparable to that of image-dependent unlearnable example methods, it does not yet surpass them. We believe this gap can be narrowed or even reversed by scaling up the training data used for the generator and further improving the model’s capacity and generalization.

\clearpage

\section*{Acknowledgments} This work is in part supported by the National Key R\&D Program of China (Grant No. 2021ZD0112804) and the National Natural Science Foundation of China (Grant No. 62276067).
The computations in this research were performed using the CFFF platform of Fudan University.

\bibliographystyle{plainnat}
\bibliography{main}

\clearpage
\appendix

\section{More Ablation Results}
\label{appendix:ablation}

\subsection{Impact of Generator Training Duration}\label{appendix:generator impact}

\begin{figure}[!ht]
	\centering
	\begin{subfigure}{0.45\linewidth}
        \includegraphics[width=\textwidth]{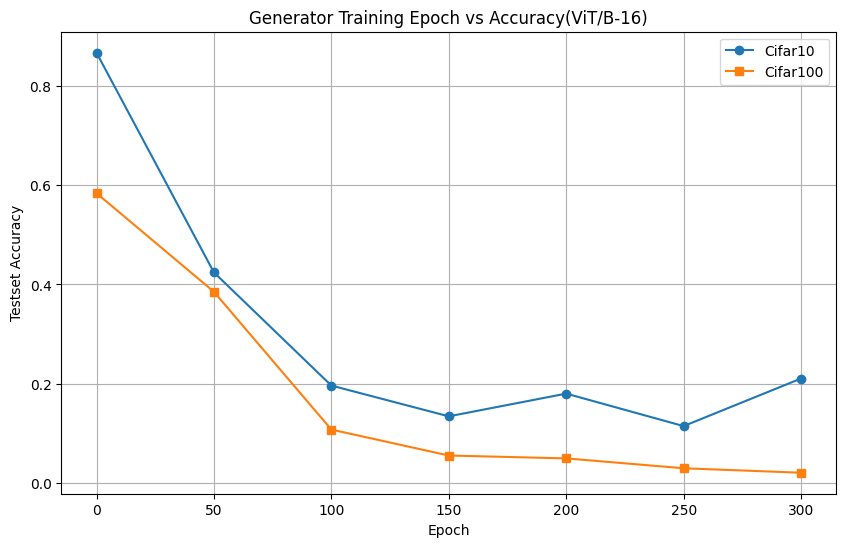} 
		\caption{ViT-B/16}
		\label{fig:vitb16_gen_acc}
	\end{subfigure}
	\begin{subfigure}{0.45\linewidth}
        \includegraphics[width=\textwidth]{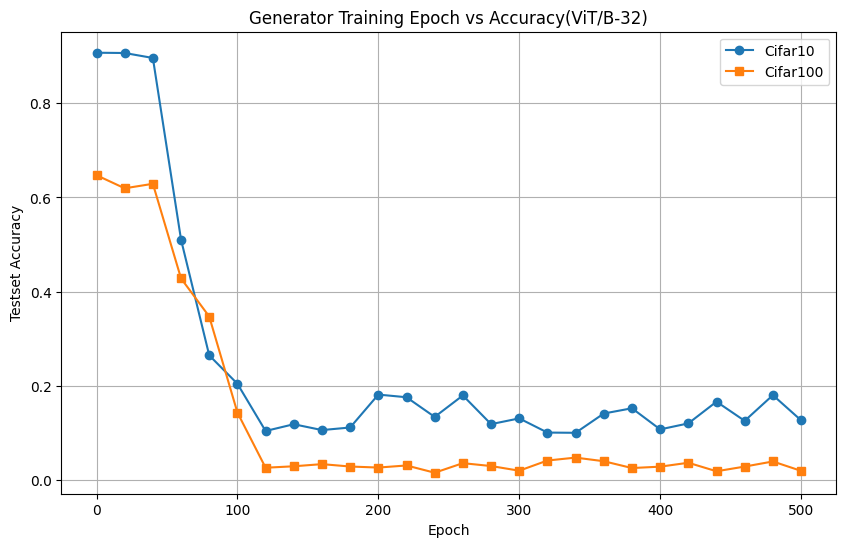} 
		\caption{ViT-B/32}
		\label{fig:vitb32_gen_acc}
	\end{subfigure}
	\caption{Acc across test set with epoch of generator training in CLIP (a) ViT-B/16 and (b) ViT-B/32}
	\label{fig:training_curve_cifar10}
\end{figure}

 We investigate how the T2UE generator's training duration influences the potency of the generated noise. Figures \ref{fig:vitb16_gen_acc} and \ref{fig:vitb32_gen_acc} plot the downstream ResNet18 test accuracy on CIFAR-10/100 using noise generated at different checkpoint epochs of the ViT-B/16 (trained 300 epochs) and ViT-B/32 (trained 500 epochs) based generators, respectively. A clear inverse relationship is observed: as the generator trains for more epochs, the resulting noise becomes significantly more effective at disrupting downstream classification, leading to substantial accuracy degradation on both CIFAR datasets. The most significant performance drop occurs relatively early in the generator's training. For instance, with the ViT-B/32 generator (Figure \ref{fig:vitb32_gen_acc}), downstream accuracy plummets within the first ~120 epochs before largely stabilizing at near-zero levels. This indicates that the generator progressively learns the mapping from text semantics to potent unlearnable perturbations, achieving strong efficacy after sufficient initial training.

\subsection{Efficiency Analysis}
\label{appendix:efficiency}

We evaluate the computational efficiency by measuring the time required to generate the unlearnable perturbations for the datasets used in our experiments in Section \ref{sec:exp_main}. The results are reported in Table \ref{tab:time_cost_unlearn}, which highlights a significant advantage for T2UE. For instance, generating noise for the entire Flickr8k dataset takes merely 0.3 hours with T2UE. This contrasts sharply with the time overhead of baseline methods: 1.1 hours for EM~\cite{huang2021unlearnable}, 4.4 hours for MEM-3~\cite{liu2024multimodal}, 10.2 hours for MEM-5~\cite{liu2024multimodal}, and the combined cost for TAP~\cite{fowl2021adversarial} which includes noise generation plus a prerequisite model training phase (4.1 hours in total). Similar efficiency gains for T2UE are observed on Flickr30k and MSCOCO. This disparity stems from the computationally intensive nature of the baselines: EM involves nested optimization loops, MEM requires complex joint optimization across modalities, and UAP necessitates costly prerequisite training. T2UE bypasses these intensive procedures by leveraging its pre-trained text-to-perturbation generator for direct noise synthesis. Consequently, T2UE offers a much more practical approach, achieving strong protection effectiveness (as shown previously) with significantly reduced computational cost.

\begin{table}[H]
\centering
\caption{Time cost for generating UEs (in hours). Note that the TAP method requires additional training of a separate model, leading to a significantly higher time overhead. In contrast, our proposed T2UE method efficiently constructs UE with the lowest time cost among all compared methods.}

\resizebox{1.0\linewidth}{!}{ 
\begin{tabular}{ccccccc}
    \toprule
    \multirow{2}{*}{Method} & \multicolumn{3}{c}{RN50} & \multicolumn{3}{c}{ViTB/32} \\
    \cmidrule(r){2-7} 
    & Flickr8K & Flickr30K & MSCOCO & Flickr8K & Flickr30K & MSCOCO \\
    \midrule[0.75pt]
    EM \cite{huang2021unlearnable}   & 1.1 & 4.4 & 2.7 & 1.1 & 5.5 & 2.3 \\
    MEM3 \cite{liu2024multimodal} & 4.4 & 36.2 & 20.4 & 8.0 & 33.3 & 18.1 \\
    MEM5 \cite{liu2024multimodal} & 10.2 & 42.4 & 21.2 & 36.2 & 37.2 & 21.2 \\
    TAP \cite{fowl2021adversarial} & 4.1 & 8.1 & 5.5 & 4.0 & 8.4 & 5.0 \\
    \midrule[0.75pt]
    \textbf{T2UE} 
         & \textbf{0.3} & \textbf{1.8} & \textbf{0.7} & \textbf{0.3} & \textbf{1.8} & \textbf{0.7} \\ 
         
    \bottomrule
\end{tabular}
}
\label{tab:time_cost_unlearn}
\end{table}

\end{document}

%% file: math_commands.tex
\usepackage{amsmath,amsfonts,bm}

\def\eqref#1{equation~\ref{#1}}

\def\1{\bm{1}}

\DeclareMathAlphabet{\mathsfit}{\encodingdefault}{\sfdefault}{m}{sl}
\SetMathAlphabet{\mathsfit}{bold}{\encodingdefault}{\sfdefault}{bx}{n}













%% file: main.bbl
\begin{thebibliography}{53}
\providecommand{\natexlab}[1]{#1}
\providecommand{\url}[1]{\texttt{#1}}
\expandafter\ifx\csname urlstyle\endcsname\relax
  \providecommand{\doi}[1]{doi: #1}\else
  \providecommand{\doi}{doi: \begingroup \urlstyle{rm}\Url}\fi

\bibitem[Birhane and Prabhu(2021)]{birhane2021large}
Abeba Birhane and Vinay~Uday Prabhu.
\newblock Large image datasets: A pyrrhic win for computer vision?
\newblock In \emph{WACV}, 2021.

\bibitem[Chen et~al.(2015)Chen, Fang, Lin, Vedantam, Gupta, Doll{\'a}r, and Zitnick]{chen2015microsoft}
Xinlei Chen, Hao Fang, Tsung-Yi Lin, Ramakrishna Vedantam, Saurabh Gupta, Piotr Doll{\'a}r, and C~Lawrence Zitnick.
\newblock Microsoft coco captions: Data collection and evaluation server.
\newblock \emph{arXiv preprint arXiv:1504.00325}, 2015.

\bibitem[Coates et~al.(2011)Coates, Ng, and Lee]{coates2011analysis}
Adam Coates, Andrew Ng, and Honglak Lee.
\newblock An analysis of single-layer networks in unsupervised feature learning.
\newblock In \emph{Proceedings of the fourteenth international conference on artificial intelligence and statistics}, 2011.

\bibitem[Cubuk et~al.(2019)Cubuk, Zoph, Mané, Vasudevan, and Le]{8953317}
Ekin~D. Cubuk, Barret Zoph, Dandelion Mané, Vijay Vasudevan, and Quoc~V. Le.
\newblock Autoaugment: Learning augmentation strategies from data.
\newblock In \emph{CVPR}, 2019.

\bibitem[DeVries and Taylor(2017)]{devries2017cutout}
Terrance DeVries and Graham~W Taylor.
\newblock Improved regularization of convolutional neural networks with cutout.
\newblock \emph{arXiv preprint arXiv:1708.04552}, 2017.

\bibitem[Dosovitskiy et~al.(2021)Dosovitskiy, Beyer, Kolesnikov, Weissenborn, Zhai, Unterthiner, Dehghani, Minderer, Heigold, Gelly, Uszkoreit, and Houlsby]{dosovitskiy2020vit}
Alexey Dosovitskiy, Lucas Beyer, Alexander Kolesnikov, Dirk Weissenborn, Xiaohua Zhai, Thomas Unterthiner, Mostafa Dehghani, Matthias Minderer, Georg Heigold, Sylvain Gelly, Jakob Uszkoreit, and Neil Houlsby.
\newblock An image is worth 16x16 words: Transformers for image recognition at scale.
\newblock \emph{ICLR}, 2021.

\bibitem[Fowl et~al.(2021)Fowl, Goldblum, Chiang, Geiping, Czaja, and Goldstein]{fowl2021adversarial}
Liam Fowl, Micah Goldblum, Ping-yeh Chiang, Jonas Geiping, Wojciech Czaja, and Tom Goldstein.
\newblock Adversarial examples make strong poisons.
\newblock \emph{NeurIPS}, 2021.

\bibitem[Fu et~al.(2022)Fu, He, Liu, Shen, and Tao]{fu2022robust}
Shaopeng Fu, Fengxiang He, Yang Liu, Li~Shen, and Dacheng Tao.
\newblock Robust unlearnable examples: Protecting data against adversarial learning.
\newblock In \emph{ICLR}, 2022.

\bibitem[Gadre et~al.(2023)Gadre, Ilharco, Fang, Hayase, Smyrnis, Nguyen, Marten, Wortsman, Ghosh, Zhang, et~al.]{gadre2023datacomp}
Samir~Yitzhak Gadre, Gabriel Ilharco, Alex Fang, Jonathan Hayase, Georgios Smyrnis, Thao Nguyen, Ryan Marten, Mitchell Wortsman, Dhruba Ghosh, Jieyu Zhang, et~al.
\newblock Datacomp: In search of the next generation of multimodal datasets.
\newblock In \emph{NeurIPS}, 2023.

\bibitem[Gong et~al.(2025)Gong, Wang, Chen, Dong, Li, Sun, Li, Wang, and Chen]{gong2025armor}
Xueluan Gong, Yuji Wang, Yanjiao Chen, Haocheng Dong, Yiming Li, Mengyuan Sun, Shuaike Li, Qian Wang, and Chen Chen.
\newblock Armor: Shielding unlearnable examples against data augmentation.
\newblock \emph{arXiv preprint arXiv:2501.08862}, 2025.

\bibitem[Guo et~al.(2023)Guo, Chen, Qiu, Guo, Luo, Chen, and Ren]{guo2023medgan}
Kehua Guo, Jie Chen, Tian Qiu, Shaojun Guo, Tao Luo, Tianyu Chen, and Sheng Ren.
\newblock Medgan: An adaptive gan approach for medical image generation.
\newblock \emph{Computers in Biology and Medicine}, 163:\penalty0 107119, 2023.

\bibitem[He et~al.(2023)He, Zha, and Katabi]{he2023indiscriminate}
Hao He, Kaiwen Zha, and Dina Katabi.
\newblock Indiscriminate poisoning attacks on unsupervised contrastive learning.
\newblock In \emph{ICLR}, 2023.

\bibitem[He et~al.(2016)He, Zhang, Ren, and Sun]{he2016deep}
Kaiming He, Xiangyu Zhang, Shaoqing Ren, and Jian Sun.
\newblock Deep residual learning for image recognition.
\newblock In \emph{CVPR}, 2016.

\bibitem[Hill(2022)]{hill2022secretive}
Kashmir Hill.
\newblock The secretive company that might end privacy as we know it.
\newblock In \emph{Ethics of Data and Analytics}, pages 170--177. Auerbach Publications, 2022.

\bibitem[Huang et~al.(2017)Huang, Liu, Van Der~Maaten, and Weinberger]{huang2017densely}
Gao Huang, Zhuang Liu, Laurens Van Der~Maaten, and Kilian~Q Weinberger.
\newblock Densely connected convolutional networks.
\newblock In \emph{Proceedings of the IEEE conference on computer vision and pattern recognition}, pages 4700--4708, 2017.

\bibitem[Huang et~al.(2021)Huang, Ma, Erfani, Bailey, and Wang]{huang2021unlearnable}
Hanxun Huang, Xingjun Ma, Sarah~Monazam Erfani, James Bailey, and Yisen Wang.
\newblock Unlearnable examples: Making personal data unexploitable.
\newblock In \emph{ICLR}, 2021.

\bibitem[Ji et~al.(2022)Ji, Ma, and Wang]{ji2022unlearnable}
Zhenlan Ji, Pingchuan Ma, and Shuai Wang.
\newblock Unlearnable examples: Protecting open-source software from unauthorized neural code learning.
\newblock In \emph{SEKE}, 2022.

\bibitem[Jia et~al.(2021)Jia, Yang, Xia, Chen, Parekh, Pham, Le, Sung, Li, and Duerig]{jia2021scaling}
Chao Jia, Yinfei Yang, Ye~Xia, Yi-Ting Chen, Zarana Parekh, Hieu Pham, Quoc Le, Yun-Hsuan Sung, Zhen Li, and Tom Duerig.
\newblock Scaling up visual and vision-language representation learning with noisy text supervision.
\newblock In \emph{ICML}, 2021.

\bibitem[Jiang et~al.(2024)Jiang, Ma, Erfani, and Bailey]{jiang2024unlearnable}
Yujing Jiang, Xingjun Ma, Sarah~Monazam Erfani, and James Bailey.
\newblock Unlearnable examples for time series.
\newblock In \emph{PAKDD}, 2024.

\bibitem[Koh and Liang(2017)]{koh2017understanding}
Pang~Wei Koh and Percy Liang.
\newblock Understanding black-box predictions via influence functions.
\newblock In \emph{ICML}. PMLR, 2017.

\bibitem[Krizhevsky(2009)]{krizhevsky2009learning}
Alex Krizhevsky.
\newblock Learning multiple layers of features from tiny images.
\newblock 2009.

\bibitem[Liao et~al.(2021)Liao, Hu, Yang, and Rosenhahn]{liao2021text}
Wentong Liao, Kai Hu, Michael~Ying Yang, and Bodo Rosenhahn.
\newblock Text to image generation with semantic-spatial aware gan.
\newblock \emph{arXiv preprint arXiv:2104.00567}, 2021.

\bibitem[Liao et~al.(2022)Liao, Hu, Yang, and Rosenhahn]{liao2022text}
Wentong Liao, Kai Hu, Michael~Ying Yang, and Bodo Rosenhahn.
\newblock Text to image generation with semantic-spatial aware gan.
\newblock In \emph{CVPR}, 2022.

\bibitem[Lin et~al.(2014)Lin, Maire, Belongie, Hays, Perona, Ramanan, Doll{\'a}r, and Zitnick]{lin2014microsoft}
Tsung-Yi Lin, Michael Maire, Serge Belongie, James Hays, Pietro Perona, Deva Ramanan, Piotr Doll{\'a}r, and C~Lawrence Zitnick.
\newblock Microsoft coco: Common objects in context.
\newblock In \emph{ECCV}, 2014.

\bibitem[Liu et~al.(2024{\natexlab{a}})Liu, Jia, Xun, Liang, and Cao]{liu2024multimodal}
Xinwei Liu, Xiaojun Jia, Yuan Xun, Siyuan Liang, and Xiaochun Cao.
\newblock Multimodal unlearnable examples: Protecting data against multimodal contrastive learning.
\newblock In \emph{MM}, 2024{\natexlab{a}}.

\bibitem[Liu et~al.(2024{\natexlab{b}})Liu, Xu, Chen, and Sun]{liu2024stable}
Yixin Liu, Kaidi Xu, Xun Chen, and Lichao Sun.
\newblock Stable unlearnable example: Enhancing the robustness of unlearnable examples via stable error-minimizing noise.
\newblock In \emph{AAAI}, volume~38, pages 3783--3791, 2024{\natexlab{b}}.

\bibitem[Longpre et~al.(2023)Longpre, Mahari, Chen, Obeng-Marnu, Sileo, Brannon, Muennighoff, Khazam, Kabbara, Perisetla, et~al.]{longpre2023data}
Shayne Longpre, Robert Mahari, Anthony Chen, Naana Obeng-Marnu, Damien Sileo, William Brannon, Niklas Muennighoff, Nathan Khazam, Jad Kabbara, Kartik Perisetla, et~al.
\newblock The data provenance initiative: A large scale audit of dataset licensing \& attribution in ai.
\newblock \emph{Nature Machine Intelligence}, 2023.

\bibitem[Ma et~al.(2025)Ma, Gao, Wang, Wang, Wang, Sun, Ding, Xu, Chen, Zhao, et~al.]{ma2025safety}
Xingjun Ma, Yifeng Gao, Yixu Wang, Ruofan Wang, Xin Wang, Ye~Sun, Yifan Ding, Hengyuan Xu, Yunhao Chen, Yunhan Zhao, et~al.
\newblock Safety at scale: A comprehensive survey of large model safety.
\newblock \emph{arXiv preprint arXiv:2502.05206}, 2025.

\bibitem[Mirza and Osindero(2014)]{2014Conditional}
Mehdi Mirza and Simon Osindero.
\newblock Conditional generative adversarial nets.
\newblock \emph{Computer Science}, pages 2672--2680, 2014.

\bibitem[Radford et~al.(2021)Radford, Kim, Hallacy, Ramesh, Goh, Agarwal, Sastry, Askell, Mishkin, Clark, et~al.]{radford2021learning}
Alec Radford, Jong~Wook Kim, Chris Hallacy, Aditya Ramesh, Gabriel Goh, Sandhini Agarwal, Girish Sastry, Amanda Askell, Pamela Mishkin, Jack Clark, et~al.
\newblock Learning transferable visual models from natural language supervision.
\newblock In \emph{ICML}, 2021.

\bibitem[Ren et~al.(2022)Ren, Xu, Wan, Ma, Sun, and Tang]{ren2022transferable}
Jie Ren, Han Xu, Yuxuan Wan, Xingjun Ma, Lichao Sun, and Jiliang Tang.
\newblock Transferable unlearnable examples.
\newblock \emph{arXiv preprint arXiv:2210.10114}, 2022.

\bibitem[Sadasivan et~al.(2023)Sadasivan, Soltanolkotabi, and Feizi]{sadasivan2023cuda}
Vinu~Sankar Sadasivan, Mahdi Soltanolkotabi, and Soheil Feizi.
\newblock Cuda: Convolution-based unlearnable datasets.
\newblock In \emph{CVPR}, 2023.

\bibitem[Sandoval-Segura et~al.(2022)Sandoval-Segura, Singla, Geiping, Goldblum, Goldstein, and Jacobs]{sandoval2022autoregressive}
Pedro Sandoval-Segura, Vasu Singla, Jonas Geiping, Micah Goldblum, Tom Goldstein, and David Jacobs.
\newblock Autoregressive perturbations for data poisoning.
\newblock \emph{NeurIPS}, 2022.

\bibitem[Sandoval-Segura et~al.(2023)Sandoval-Segura, Singla, Geiping, Goldblum, and Goldstein]{sandoval2023can}
Pedro Sandoval-Segura, Vasu Singla, Jonas Geiping, Micah Goldblum, and Tom Goldstein.
\newblock What can we learn from unlearnable datasets?
\newblock \emph{NeurIPS}, 2023.

\bibitem[Schuhmann et~al.(2021)Schuhmann, Vencu, Beaumont, Kaczmarczyk, Mullis, Katta, Coombes, Jitsev, and Komatsuzaki]{schuhmann2021laion}
Christoph Schuhmann, Richard Vencu, Romain Beaumont, Robert Kaczmarczyk, Clayton Mullis, Aarush Katta, Theo Coombes, Jenia Jitsev, and Aran Komatsuzaki.
\newblock Laion-400m: Open dataset of clip-filtered 400 million image-text pairs.
\newblock \emph{arXiv preprint arXiv:2111.02114}, 2021.

\bibitem[Schuhmann et~al.(2022)Schuhmann, Beaumont, Vencu, Gordon, Wightman, Cherti, Coombes, Katta, Mullis, Wortsman, et~al.]{schuhmann2022laion}
Christoph Schuhmann, Romain Beaumont, Richard Vencu, Cade Gordon, Ross Wightman, Mehdi Cherti, Theo Coombes, Aarush Katta, Clayton Mullis, Mitchell Wortsman, et~al.
\newblock Laion-5b: An open large-scale dataset for training next generation image-text models.
\newblock \emph{NeurIPS}, 2022.

\bibitem[Simonyan and Zisserman(2014)]{simonyan2014very}
Karen Simonyan and Andrew Zisserman.
\newblock Very deep convolutional networks for large-scale image recognition.
\newblock \emph{arXiv preprint arXiv:1409.1556}, 2014.

\bibitem[Sun et~al.(2024)Sun, Zhang, Zhang, Ma, and Jiang]{sun2024unseg}
Ye~Sun, Hao Zhang, Tiehua Zhang, Xingjun Ma, and Yu-Gang Jiang.
\newblock Unseg: One universal unlearnable example generator is enough against all image segmentation.
\newblock In \emph{NeurIPS}, 2024.

\bibitem[Szegedy et~al.(2015)Szegedy, Liu, Jia, Sermanet, Reed, Anguelov, Erhan, Vanhoucke, and Rabinovich]{szegedy2015going}
Christian Szegedy, Wei Liu, Yangqing Jia, Pierre Sermanet, Scott Reed, Dragomir Anguelov, Dumitru Erhan, Vincent Vanhoucke, and Andrew Rabinovich.
\newblock Going deeper with convolutions.
\newblock In \emph{CVPR}, 2015.

\bibitem[Vondrick et~al.(2016)Vondrick, Pirsiavash, and Torralba]{vondrick2016generating}
Carl Vondrick, Hamed Pirsiavash, and Antonio Torralba.
\newblock Generating videos with scene dynamics.
\newblock \emph{NeurIPS}, 29, 2016.

\bibitem[Wang et~al.(2024{\natexlab{a}})Wang, Xue, Li, Camtepe, and Zhu]{wang2024provably}
Derui Wang, Minhui Xue, Bo~Li, Seyit Camtepe, and Liming Zhu.
\newblock Provably unlearnable data examples.
\newblock \emph{arXiv preprint arXiv:2405.03316}, 2024{\natexlab{a}}.

\bibitem[Wang et~al.(2022)Wang, Yang, Hu, Li, Lin, Gan, Liu, Liu, and Wang]{wang2022git}
Jianfeng Wang, Zhengyuan Yang, Xiaowei Hu, Linjie Li, Kevin Lin, Zhe Gan, Zicheng Liu, Ce~Liu, and Lijuan Wang.
\newblock Git: A generative image-to-text transformer for vision and language.
\newblock \emph{arXiv preprint arXiv:2205.14100}, 2022.

\bibitem[Wang et~al.(2024{\natexlab{b}})Wang, Li, Liu, Zhang, Hu, Zhang, Zhou, and Jin]{wang2024unlearnable}
Xianlong Wang, Minghui Li, Wei Liu, Hangtao Zhang, Shengshan Hu, Yechao Zhang, Ziqi Zhou, and Hai Jin.
\newblock Unlearnable 3d point clouds: Class-wise transformation is all you need.
\newblock In \emph{NeurIPS}, 2024{\natexlab{b}}.

\bibitem[Wu et~al.(2020)Wu, Chen, and Meng]{wu2020dcgan}
Qiufeng Wu, Yiping Chen, and Jun Meng.
\newblock Dcgan-based data augmentation for tomato leaf disease identification.
\newblock \emph{IEEE access}, 8:\penalty0 98716--98728, 2020.

\bibitem[Xu et~al.(2024)Xu, Xie, Tan, Huang, Howes, Sharma, Li, Ghosh, Zettlemoyer, and Feichtenhofer]{xu2024demystifying}
Hu~Xu, Saining Xie, Xiaoqing Tan, Po-Yao Huang, Russell Howes, Vasu Sharma, Shang-Wen Li, Gargi Ghosh, Luke Zettlemoyer, and Christoph Feichtenhofer.
\newblock Demystifying clip data.
\newblock In \emph{ICLR}, 2024.

\bibitem[Yang et~al.(2024)Yang, Gao, and Mirzasoleiman]{yang2024robust}
Wenhan Yang, Jingdong Gao, and Baharan Mirzasoleiman.
\newblock Robust contrastive language-image pretraining against data poisoning and backdoor attacks.
\newblock \emph{NeurIPS}, 2024.

\bibitem[Yang et~al.(2023)Yang, He, Li, Backes, Humbert, Berrang, and Zhang]{yang2023data}
Ziqing Yang, Xinlei He, Zheng Li, Michael Backes, Mathias Humbert, Pascal Berrang, and Yang Zhang.
\newblock Data poisoning attacks against multimodal encoders.
\newblock In \emph{ICML}. PMLR, 2023.

\bibitem[Ye et~al.(2025)Ye, Su, and Qian]{ye2025how}
Kai Ye, Liangcai Su, and Chenxiong Qian.
\newblock How far are we from true unlearnability?
\newblock In \emph{ICLR}, 2025.

\bibitem[Young et~al.(2014)Young, Lai, Hodosh, and Hockenmaier]{young2014image}
Peter Young, Alice Lai, Micah Hodosh, and Julia Hockenmaier.
\newblock From image descriptions to visual denotations: New similarity metrics for semantic inference over event descriptions.
\newblock \emph{Transactions of the Association for Computational Linguistics}, 2:\penalty0 67--78, 2014.

\bibitem[Yu et~al.(2022)Yu, Zhang, Chen, Yin, and Liu]{yu2022availability}
Da~Yu, Huishuai Zhang, Wei Chen, Jian Yin, and Tie-Yan Liu.
\newblock Availability attacks create shortcuts.
\newblock In \emph{SIGKDD}, 2022.

\bibitem[Zhang et~al.(2017)Zhang, Xu, Li, Zhang, Wang, Huang, and Metaxas]{zhang2017stackgan}
Han Zhang, Tao Xu, Hongsheng Li, Shaoting Zhang, Xiaogang Wang, Xiaolei Huang, and Dimitris~N Metaxas.
\newblock Stackgan: Text to photo-realistic image synthesis with stacked generative adversarial networks.
\newblock In \emph{ICCV}, 2017.

\bibitem[Zhang(2017)]{zhang2017mixup}
Hongyi Zhang.
\newblock mixup: Beyond empirical risk minimization.
\newblock \emph{arXiv preprint arXiv:1710.09412}, 2017.

\bibitem[Zhang et~al.(2023)Zhang, Ma, Yi, Sang, Jiang, Wang, and Xu]{zhang2023unlearnable}
Jiaming Zhang, Xingjun Ma, Qi~Yi, Jitao Sang, Yu-Gang Jiang, Yaowei Wang, and Changsheng Xu.
\newblock Unlearnable clusters: Towards label-agnostic unlearnable examples.
\newblock In \emph{CVPR}, 2023.

\end{thebibliography}
